\documentclass[letterpaper, 10 pt, conference]{ieeeconf}  

\IEEEoverridecommandlockouts                              

\usepackage{tabu}
\usepackage{graphicx}
\usepackage{newclude}
\usepackage{color}
\usepackage[dvipsnames, svgnames, x11names]{xcolor}
\usepackage[colorlinks,linkcolor=blue,breaklinks=true]{hyperref}
\usepackage[para,online,flushleft]{threeparttable}
\usepackage{etoolbox}

\usepackage{stfloats}
\usepackage{subcaption}
\usepackage{amsmath}
\usepackage{ulem}
\usepackage{cite}
\usepackage{url}
\usepackage{verbatim}
\usepackage{booktabs}
\usepackage{pifont}
\usepackage{algorithm}
\usepackage{algpseudocode}
\usepackage{dsfont}
\usepackage{amsfonts} 
\usepackage{setspace}
\usepackage[skip=1.5pt]{caption}
\usepackage{tikz,pgfplots,epsfig}
\usepackage{balance}
\pgfplotsset{compat=1.17}
\usetikzlibrary{positioning}
\usetikzlibrary{matrix}
\usetikzlibrary{shapes.multipart} 
\usetikzlibrary{arrows.meta}
\makeatletter
\patchcmd{\@makecaption}
  {\scshape}
  {}
  {}
  {}
\makeatletter
\patchcmd{\@makecaption}
  {\\}
  {.\ }
  {}
  {}
\makeatother

\title{\LARGE \bf
Inference-stage Adaptation-projection Strategy Adapts \\ Diffusion Policy to Cross-manipulators Scenarios}

\author{Xiangtong Yao$^{1}$, Yirui Zhou$^{1}$, Yuan Meng$^{1}$, Yanwen Liu$^{1}$, Liangyu Dong$^{1}$, \\ Zitao Zhang$^{1}$, Zhenshan Bing$^{1,2,\dagger}$, Kai Huang$^{3}$, Fuchun Sun$^{4}$, Alois Knoll$^{1}$
\thanks{$^1$ Technical University of Munich, Munich, Germany}
\thanks{$^2$ Nanjing University, Nanjing, China}
\thanks{$^3$ Sun Yat-sen University, Guangzhou, China}
\thanks{$^4$ Tsinghua University, Beijing, China}
\thanks{$^{\dagger}$Corresponding author: Zhenshan Bing {\tt\small zhenshan.bing@tum.de}}}

\providecommand{\keywords}[1]{\textbf{\textit{Keywords---}} #1}
\usepackage{pgfplots}
\usepackage{pgfplotstable}
\usepackage{tikz}
\tikzset{>=stealth}
\usetikzlibrary{patterns}
\usetikzlibrary{pgfplots.statistics}
\usetikzlibrary{positioning, shapes}
\usetikzlibrary{calc}
\usetikzlibrary{backgrounds,scopes}
\usetikzlibrary{fit}
\usetikzlibrary{arrows.meta}
\tikzstyle{arrow} = [thick,->,>=stealth,rounded corners=4pt, draw=black, align=center]
\tikzstyle{arrow1} = [line width=0.2mm,->,>=stealth,rounded corners=4pt, draw=black!70, align=center]
\tikzstyle{origin} = [circle,fill=black,thick,inner sep=0pt, minimum size=1mm]
\tikzstyle{target1} = [circle,fill=green,thick,inner sep=0pt, minimum size=1mm]
\tikzstyle{target2} = [circle,fill=orange,thick,inner sep=0pt, minimum size=1mm]
\tikzstyle{target3} = [circle,fill=red,thick,inner sep=0pt, minimum size=1mm]
\tikzstyle{endpoint} = [circle,fill=black,thick,inner sep=0pt, minimum size=0.5mm]
\tikzstyle{arcstyle} = [start angle=0, end angle=180, radius=10mm]
\tikzstyle{arrow2} = [thick,<->,>=stealth,rounded corners=4pt, draw=black, align=center]

\definecolor{color-obs-seq}{RGB}{219, 232, 213}
\definecolor{color-act-seq}{RGB}{172, 204, 255}
\definecolor{color-opti}{RGB}{248,206,204} 
\definecolor{color-gmap}{RGB}{0,204,153} 

\definecolor{ffblue}{RGB}{097, 108, 140}
\definecolor{ffdarkgreen}{RGB}{086, 140, 135}
\definecolor{fflightgreen}{RGB}{178, 213, 155}
\definecolor{ffyellow}{RGB}{242, 222, 121}
\definecolor{ffred}{RGB}{217, 095, 024}
\definecolor{ffred_pv}{RGB}{202, 074, 046}
\definecolor{fforange_pv}{RGB}{232, 141, 047}
\definecolor{ffgreen_pv}{RGB}{059, 165, 149}
\definecolor{ffgreendark_pv}{RGB}{032, 117, 106}
\definecolor{nature_tab_gray1}{HTML}{D8D6C2}
\definecolor{nature_tab_gray2}{HTML}{ECEADF}

\definecolor{frankacolor}{HTML}{2CA02C}

\definecolor{forestgreen}{rgb}{0.13, 0.55, 0.13}

\definecolor{graspw}{RGB}{0, 128, 0}
\definecolor{dp}{RGB}{64, 224, 208}
\definecolor{dp3}{RGB}{63, 63, 255}
\definecolor{ours}{RGB}{148, 0, 211}
\definecolor{blockcolor}{RGB}{238, 130, 238}
\definecolor{safeline}{RGB}{255, 0, 0}
\definecolor{bananacolor}{RGB}{255, 165, 0}

\let\oldtwocolumn\twocolumn
\renewcommand\twocolumn[1][]{%
    \oldtwocolumn[{#1}{
    \vspace{-5pt}
    \begin{center}
          \textcolor{orange}{This work has been submitted to the IEEE for possible publication.}

          \textcolor{orange}{Copyright may be transferred without notice, after which this version may no longer be accessible.}

          \vspace{5pt}
          \includegraphics[width=\textwidth]{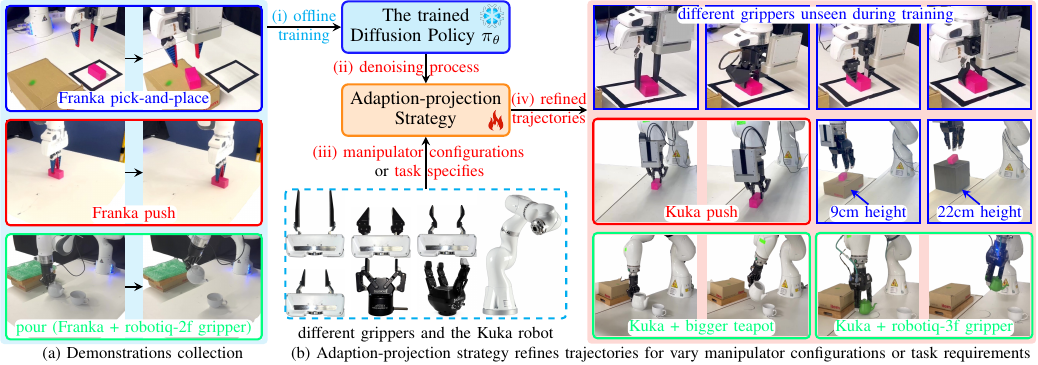}
          \captionof{figure}{(a) Collect demonstrations via the base manipulator (Franka robot with a specific gripper) to train the diffusion policy $\pi_\theta$ for each task. (b) Our adaptation-projection strategy enables zero-shot adaptation of the policy to new manipulator configurations (Kuka robot, different grippers) and task requirements (e.g., obstacle heights) at inference time.} 
           \label{fig:first-figure}
    \end{center}
    }]
}

\begin{document}

\maketitle



\begin{abstract}

Diffusion policies are powerful visuomotor models for robotic manipulation, yet they often fail to generalize to manipulators or end-effectors unseen during training and struggle to accommodate new task requirements at inference time. Addressing this typically requires costly data recollection and policy retraining for each new hardware or task configuration. To overcome this, we introduce an adaptation-projection strategy that enables a diffusion policy to perform zero-shot adaptation to novel manipulators and dynamic task settings, entirely at inference time and without any retraining. Our method first trains a diffusion policy in SE(3) space using demonstrations from a base manipulator. During online deployment, it projects the policy's generated trajectories to satisfy the kinematic and task-specific constraints imposed by the new hardware and objectives. Moreover, this projection dynamically adapts to physical differences (e.g., tool-center-point offsets, jaw widths) and task requirements (e.g., obstacle heights), ensuring robust and successful execution. We validate our approach on real-world pick-and-place, pushing, and pouring tasks across multiple manipulators, including the Franka Panda and Kuka iiwa 14, equipped with a diverse array of end-effectors like flexible grippers, Robotiq 2F/3F grippers, and various 3D-printed designs. Our results demonstrate consistently high success rates in these cross-manipulator scenarios, proving the effectiveness and practicality of our adaptation-projection strategy. The code will be released after peer review.

\end{abstract}

\section{Introduction}
Deep imitation learning has shown promise in learning manipulation skills from human demonstrations~\cite{10685120,di2023one}, reducing the need for extensive programming efforts compared to rule-based approaches (i.e., motion planning~\cite{migimatsu2020object}) and reinforcement learning~\cite{10341769,10160626,meng2025preserving}. Among these, diffusion policies~\cite{chi2023diffusion,Ze2024DP3} integrate diffusion model~\cite{ho2020denoising} into imitation learning scheme and demonstrate powerful capabilities in learning visumotor manipulation skills, generating diverse and adaptable strategies in varying scenarios, such as multi-task manipulations\cite{ma2024hierarchical}, one-shot learning \cite{xue2025demogen}. However, these methods typically assume fixed manipulator configurations (robot and end-effector) throughout training and deployment, struggling to adapt to new manipulator configurations. For instance, when switching from a base gripper to one with a different morphology, the robot's tool-center-point (TCP) may shift (as illustrated in different grippers in Fig. \ref{fig:first-figure}), leading to potential collisions or missed grasps if the pretrained policy is naively applied to generate trajectories. 

Recent research mitigates this limitation through multi-embodiment learning strategies\cite{yang2023polybot,wang2024cross,wang2024scaling,chen2024mirage,bauer2025latent,chen2025dexonomy,seo2025legato,freiberg2025diffusion}. However, these methods often require collecting substantial embodiment-specific data\cite{wang2024scaling,freiberg2025diffusion}, relying on policy fine-tuning to accommodate new configurations\cite{wang2024scaling,wang2024cross}, or requiring united task distributions during training and developing \cite{chen2024mirage} synthesis of large-scale heterogeneous gripper datasets\cite{chen2025dexonomy}, training an united policy across multiple manipulators\cite{yang2023polybot,freiberg2025diffusion,bauer2025latent}, designing a new grasping tool for across-embodiments scenarios\cite{seo2025legato}. When adapting the trained policy to new manipulator configurations (including robot and end-effector that were unseen during policy training) to satisfy different task requirements (like picking a heavier object), the above approaches incur time-consuming adaptation costs and limit deployment flexibility in real-world scenarios.

We bridge this gap through an integration of diffusion policy~\cite{chi2023diffusion,Ze2024DP3} with a adaption-projection strategy. This strategy allows the trained policy to fit new manipulator configurations without policy retraining or fine-tuning, as shown in Fig.\ref{fig:first-figure}. The policy first learns manipulation primitives following diffusion policy, where the training demonstration data is collected via a base configuration in Fig.\ref{fig:first-figure}\textcolor{blue}{(a)}. During the policy's inference stage, the base configuration is replaced with a new one (different grippers or Kuka robot). To ensure the generative trajectories are valid for new configurations, we first adapt the manipulator-specific dimensional offsets to the observation inputs of the policy, and recast the traditional denoising process as a constrained-optimization projection process. The projection introduces mathematical encoding of task requirements and safety constraints (e.g., object grasping, avoiding collisions) to the denoising process, utilizing a quadratic programming optimization method to incrementally refine an initially noisy trajectory to conform to unseen manipulators and new task requirements (like placing the object into a higher platform), as shown in Fig.\ref{fig:first-figure}\textcolor{blue}{(b)}. Crucially, this approach does not require policy retraining or fine-turning, preserving flexibility for real-world deployment with minimal overhead. Key contributions are summarized as follows.
\begin{itemize}
    \item A adaptation-projection strategy is introduced into the inference-stage of the trained diffusion policy, adjusting and refining the generation of trajectories to counteract the TCP offset problem caused by switching manipulator configurations in SE(3) space, ensuring task completion and the satisfaction of new task requirements without retraining and fine-tuning the policy. 
    \item A general formulation of the task and safety constraints is designed for the adaptation-projection process. Moreover, a cumulative trajectory refinement process is utilized to preserve the trajectory's temporal consistency. 
    \item Real-world experiments for varying-difficulty manipulation tasks (pick-and-place, pushing, and pouring tea) in across-grippers and across-manipulator (Franka, Kuka) scenarios show the effectiveness of our method.
\end{itemize} 

\section{Related Work}

\textbf{Cross-manipulator Skill Transfer:} Transferring robot manipulation knowledge between different embodiments can improve the policy's flexibility and generalizability for adapting to new tasks or hardware settings~\cite{wang2024scaling}, including the knowledge transitions between different grippers~\cite{zakka2022xirl,freiberg2025diffusion} or robot manipulators~\cite{yang2023polybot,tatiya2023transferring,wang2024scaling,wang2024cross,seo2025legato}. For example, Zakka et al.~\cite{zakka2022xirl} combine the temporal cycle-consistency method with imitation learning to learn an invariant feature space for different embodiments, empowering target robot manipulators to perform the tasks by imitating video demonstrations of human experts. While the feature correspondence between the source and target domains has to be re-trained if a new source or target domain is introduced. To mitigate this limitation, a dominant approach involves learning aligned latent feature spaces between source and target robots using techniques like adversarial training and cycle consistency~\cite{wang2024cross, tatiya2023transferring}. By leveraging large-scale heterogeneous robot data, Wang et al.~\cite{wang2024scaling} propose a heterogeneous pre-trained transformer that can perform different tasks across heterogeneous robot manipulators. However, these approaches relies on transfer learning or domain adaption~\cite{wang2024cross,wang2024scaling,tatiya2023transferring}, where introducing new hardware configuration (e.g., grippers, manipulators) to the policy require fine-tuning policy~\cite{wang2024scaling} or retraining auxiliary networks~\cite{wang2024cross,tatiya2023transferring}, limiting the flexibility of the policy. Another line of work, such as Mirage~\cite{chen2024mirage}, uses visual inpainting to transfer policies to unseen robots. This approach, however, struggles with significant changes in background or large tool-center-point (TCP) shifts between robots. Yao et al.~\cite{yao2025pick} propose a learning-optimization diffusion policy that adapts to new grippers without retraining, but their method is limited to translational TCP shifts and is not validated in cross-manipulator scenarios. Our method circumvents these limitations by integrating an adaptation-projection strategy into the inference stage of the diffusion policy~\cite{chi2023diffusion} that requires no retraining or fine-tuning. The adaptation component explicitly handles geometric discrepancies like TCP shifts (larger than 10cm shifts in translational axis and shift issue in rotational axes), while the projection refines the generated trajectory online to ensure it complies with the safety and task constraints of the new hardware in cross-manipulator scenarios. This enables a zero-shot adaptation to manipulators with new configurations.

\textbf{Inference-stage Refining Trajectory in Diffusion Policy:} Diffusion models have shown promise for solving decision-making tasks, including motion planning~\cite{10342382,mizuta2024cobl}, imitation learning~\cite{chi2023diffusion,Ze2024DP3,reuss2023goal}, and reinforcement learning~\cite{liang2023adaptdiffuser,venkatraman2024reasoning}. When solving decision-making tasks through generative models, the policy needs not only to accomplish the task goal but also to satisfy certain task constraints, such as collision avoidance. Some works add a residual loss to the training objectives if the task constraints are consistent during the policy training and inference~\cite{mizuta2024cobl,bastek2024physics}. A more flexible method is training diffusion models via classifier-free guidance\cite{ho2022classifier}, introducing conditioning variables that represent constraints into the policy training, such as physical constraints for guiding human motion generation~\cite{yuan2023physdiff}. However, the model conditioning encourages the generated samples to adhere to task constraints rather than strict guarantee constraints~\cite{christopher2024constrained}. Alternative post-processing methods draw constraints into the last denoising stage of the sample-generating process and obtain samples that satisfy the constraints by solving an optimization problem~\cite{maze2023diffusion,giannone2023aligning}. As the optimization problem does not consider the unknown data likelihood, post-processing may result in samples significantly deviating from the data distribution~\cite{romer2024diffusion}. To mitigate this issue, an iterative projection is integrated into the denoising process, confining each generated sample to the constraints. Some works employ it for sequential decision-making~\cite{christopher2024constrained}, which is time-consuming. Or model-based trajectory control~\cite{romer2024diffusion}, falling short of seamlessly adapting to a new model since introducing a new gripper changes the robot dynamics. Our approach utilizes adaptive projection alongside optimization techniques in the denoising process but mitigates the time consumption and makes the policy seamlessly adaptable to new manipulator configurations and task requirements (i.e., place the object to a higher platform).

\section{Methodology}
\subsection{Problem Statement}

We formalize the challenge of gripper-agnostic manipulation via diffusion policies through three core components:

\input{figure-exp-gripper-v2}
\input{figure-pour-gripper-v2}

\begin{itemize}
    \item \textbf{Gripper Configuration}: Let $\mathbb{G} \subset \mathbb{R}^{d_g}$ denote the space of parallel gripper parameters encoding maximum width $w^{\max}$ and gripper height $h^{\text{}}$, as shown in Fig. \ref{fig:grippers}. 
    \item \textbf{Observation Domain}: $\mathcal{O} = \mathcal{S}_{\text{sce}} \times \mathcal{S}_{\text{rob}}$ where $\mathcal{S}_{\text{sce}}$ represents scene observations (3D point clouds) and $\mathcal{S}_{\text{rob}} = SE(3) \times [0,g]$ the robot state (end-effector pose $\mathbf{x}_{\text{ee}} \in SE(3)$, gripper width $g$) $\mathbf{x}_{\text{ee}}$ is gripper-agnostic and different across robots, as shown in Fig. \ref{fig:pour-grippers}. $g$ is specific for grippers.
    \item \textbf{Action Space}: $\mathcal{A} \subset \mathbb{R}^{d_a}$ containing end-effector displacements $\Delta\mathbf{x}_{\text{ee}}$ and gripper commands. 
\end{itemize}

The policy $\pi_\theta$ is trained on demonstrations $\mathcal{D} = \{\tau^{(i)}\}_{i=1}^N$ collected with a base configuration, such as Franka with the gripper $\mathbb{G}_0 \in \mathbb{G}$. Each trajectory $\tau = \{(\mathbf{o}_t, \mathbf{a}_t)\}_{t=0}^T$ satisfies: $\mathbf{o}_t = (\mathcal{S}^0_{\text{sce}}, \mathbf{x}_{\text{ee}}^0, g_t^0)$ and $\mathbf{a}_t \sim \pi_{\text{expert}}(\cdot|\mathbf{o}_t)$,

where superscript $0$ indicates $\mathbb{G}_0$ parameters. During deployment with new configuration (Kuka + gripper $\mathbb{G}_i \neq \mathbb{G}_0$), the observation-action distribution shifts due to (1) visual/kinematic differences $\mathcal{O}^i \neq \mathcal{O}^0$ (EE base differs in different robots and gripper kinematics variations result in tool-center-point (TCP) shifts in Fig. \ref{fig:pour-grippers}), and (2) policy mismatch $p_{\theta}(\mathcal{A}|\mathcal{O}^i) \neq p_{\theta}(\mathcal{A}|\mathcal{O}^0)$. This manifests as trajectory divergence $\|\tau_{1:T}^{\mathbb{G}_i} - \tau_{1:T}^{\mathbb{G}_0}\|_{\mathcal{W}} > \delta_{\text{tol}}$, where $\mathcal{W}$ is the task-specific metric and $\delta_{\text{tol}}$ the success threshold, e.g., objects cannot be grasped with shorter grippers, and collisions can result from using longer grippers.

To mitigate these issues, we develop policy $\pi_\theta^*$ that maintains task performance under manipulator variation, combining (1) Adaptation of manipulator configurations for trajectory alignment and (2) Task-satisfied and safety-constrained trajectory projection in the policy inference stage without retraining the policy. The overview framework is shown in Fig. \ref{fig:framework}.

\input{framework-figure2-v2}

\subsection{Refining Trajectory Generation} 
\label{sec:optimization}
The training scheme of the base policy $\pi_{\theta}$ is consistent with that of Diffusion Policy~\cite{chi2023diffusion} (DDPM) with observations $\mathcal{O}$ and MSE training loss. During inference, $\pi_{\theta}$ employs Denoising Diffusion Implicit Models (DDIM)\cite{song2021denoising} to generate trajectories. In the denoising process, however, our method introduces an adaptation-projection strategy into DDIM to enforce the generative trajectory to fit different manipulator configurations, ensuring task completion and robot safety. 

\textbf{Manipulator configuration adaptation:} As shown in Fig. \ref{fig:pour-grippers}, during identical object manipulation, end-effector base (EE base) variations of different robots and gripper variations induce grasping width differences in the gripper state ($g$) and the tool-center-point (TCP) discrepancies, primarily along: (1) vertical axis ($z$): TCP translational offset ($\|d_1 - d_2\|$), (2) rotational axes ($\theta_x$, $\theta_y$): TCP rotational offset. These discrepancies cause inconsistent actions predicted by $\pi_\theta$ across grippers or failure of solving task when the robot perform unadapted trajectories (such as pouring in Fig. \ref{fig:pour-grippers}(c)).

We define mapping expressions that project $\mathbb{G}_{(i)}$ parameters to the $\mathbb{G}_{(0)}$ basis, where $\mathbb{G}_{(0)}$ denotes the base gripper, and $\mathbb{G}_{(i)}$ represent a new gripper of category $i$: 
\begin{equation}
\label{eq:gripper_mapping}
\begin{aligned}
    z'_{(i)}  &= z_{(i)} + \Delta d_{(i)}, \\
    g'_{(i)} &= g_{(0)}^{\rm max} - \alpha_{(i)} (g_{(i)}^{\rm max} - g_{(i)}),
\end{aligned}
\end{equation}
where $z_{(i)}$ is the measured height of EE base when the end-effector equipped with $\mathbb{G}_{(i)}$, $\Delta d_{(i)} = z_{(0)} - z_{(i)}$ is the TCP offset from the base manipulator configuration, $g_{(i)} \in [g^{\min}_{(i)}, g^{\max}_{(i)}]$ is the real-time grasping width, $\alpha_{(i)} = (g^{\max}_{(0)}-g^{\rm grasp}_{(0)})/(g^{\max}_{(i)}-g^{\rm grasp}_{(i)})$ scales widths, $g^{\rm grasp}$ is the width when the gripper grasps the object. The mapping parameters $\{\Delta h_{(i)}, \alpha_{(i)}\}$ are obtained through offline calibration with mechanical measurement of gripper dimensions and EE base offsets across different robots. 

For the adaptation of rotational axes (taking $\theta_x$ as an example, the same applies to $\theta_y$), the key idea is let $d_1\sin\theta_x = d_2\sin\theta^*_x$, such that $l_1 = l^*_2$, as shown in Fig. \ref{fig:pour-grippers}(d), the final adapted angle $\theta^*_x$ is:
\begin{equation}
\label{eq:rotational_mapping}
    \theta^*_x = \arcsin\left(\tfrac{d_1}{d_2}\sin\theta_x\right),
\end{equation}
during task performing, to achieve the adapted state $\theta^*_x$, the process is: (1) input the current adapted state $\theta^1_{t0}=\arcsin(\frac{d_2\sin\theta^2_{t0}}{d_1})$ to the policy $\pi_\theta$, obtaining the  action $\Delta\theta^1_{t0}$ for the robot 1. (2) The next expected state $\theta^1_{t1}=\theta^1_{t0}+\Delta\theta^1_{t0}$. (3) The adapted action $\Delta \theta^2_{t0}=\arcsin(\frac{d_1\sin\theta^1_{t1}}{d_2})-\theta^2_{t0}$ for the robot 2. (4) The robot 2 executes the action $\Delta \theta^2_{t0}$. Reply (1)-(4) until the adapted state $\theta^*_x$ is reached in the robot 2.

During policy execution, the transformed pose $\mathcal{S}'_{\text{rob}} = (x,y,z'_{(i)},\theta'_{(i)},g'_{(i)})$ is fed to $\pi_\theta$ instead of $\mathcal{S}_{\text{rob}}$ in observations $\mathcal{O}$, maintaining trajectory consistency across grippers and manipulators.

\textbf{Task-satisfied and Safety-constrained Trajectory Projection} 
While the configuration adaptation of different grippers aligns geometric parameters, visual perception differences from gripper morphology can still induce unsafe trajectory variations. To guarantee constraint satisfaction, we integrate a projection layer into the DDIM denoising process~\cite{song2021denoising}. The modified reverse diffusion step becomes:
\begin{equation}\label{eq:modified_ddim}
  \mathbf{a}^{k-1}_{t} = \text{Proj}_{\mathcal{C}}\left(\mu_{k}(\mathbf{a}^k_{t}, \epsilon_{\theta}(\mathbf{a}^k_{t},\mathbf{o}_{t},k))\right),
\end{equation}
where $\text{Proj}_{\mathcal{C}}(\cdot)$ enforces safety constraints $\mathcal{C}$ through the following two steps.

\noindent\paragraph{\textbf{Constraint-aware denoising}}
For efficiency, projection activates only in the final denoising steps ($k \leq 5$). At each step $k$, we solve a quadratic program problem:
\vspace{-0.5em}
\begin{subequations}
\begin{align}
    \nu^{k*}_t &= \underset{\nu^k_t}{\arg\min} \|\nu^k_t\|^2_2 \nonumber\\
    \text{s.t. } & \|\mathcal{S}'_{\text{rob}}(z)_t + \Phi\left(\mathbf{a}^{k}_t\right) + \nu^k_t\|_{1} \geq \epsilon_{\text{safe}}, \label{eq:safety_optimization-1}\\
               & \|\arcsin\left[\tfrac{d_1}{d_2}\sin\left(\mathcal{S}'_{\text{rob}}(\theta)_t+\Phi(\mathbf{a}^{k}_t)\right)\right] - \label{eq:safety_optimization-2}\\
               &\qquad \arcsin\left[\tfrac{d_1}{d_2}\sin(\mathcal{S}'_{\text{rob}}(\theta)_t)\right] - \nu^k_t\|_{1} \leq \epsilon_{\text{task}}, \nonumber
\end{align}
\end{subequations}
where $\Phi(\cdot)$ maps latent actions to Cartesian displacement, which is denormalization in our case, $\epsilon_{\text{safe}} = 0.01$ m (safety margin), $\epsilon_{\text{task}} = 0.05$ rad (task margin), and $\nu^k_t\in \mathbb{R}^6$ is the corrective offset for translational and rotational movements.

\noindent\paragraph{\textbf{Temporal consistency enforcement}} 
To maintain safety over the policy's $T_a$-step action horizon ($j \in [0,T_a-1]$), we extend \eqref{eq:safety_optimization-1} and \eqref{eq:safety_optimization-2} with cumulative constraints:
\begin{equation}\label{eq:sequence_constraint}
\|\mathcal{S}'_{\text{rob}}(z)_{t} + \sum_{r=0}^{j}\Phi(\mathbf{a}^{k}_{t+r}) + \nu^k_{t+j}\|_{1} \geq \epsilon_{\text{safe}}
\end{equation}
\vspace{-0.8em}
\begin{equation}\label{eq:sequence_constraint-2}
    \begin{aligned}
     &\|\arcsin\left[\tfrac{d_1}{d_2}\sin\left(\mathcal{S}'_{\text{rob}}(\theta)_t+\sum_{r=0}^{j}\Phi(\mathbf{a}^{k}_{t+r})\right)\right] - \\
               &\qquad \arcsin\left[\tfrac{d_1}{d_2}\sin(\mathcal{S}'_{\text{rob}}(\theta)_t)\right] - \nu^k_{t+j}\|_{1} \leq \epsilon_{\text{task}}
    \end{aligned}
\end{equation}

\begin{algorithm}[!t]
    \caption{Cross-manipulator Trajectory Generation}\label{alg:safe_diffusion}
    \setstretch{1}
    \begin{algorithmic}[1]
    
        \Require Novel gripper $\mathbb{G}_i$, Observation $\mathbf{o}_t$, trained noise predicted $\epsilon_\theta$ \textcolor{gray}{(\textbf{Inference Stage})}
        \Ensure Safe and task-specified trajectory $\tau = \{\mathbf{a}_{t:t+T_a-1}\}$
        
        \State $\mathcal{S}'_{\text{rob}} \leftarrow \text{Equation}\ \eqref{eq:gripper_mapping},\eqref{eq:rotational_mapping} \textcolor{gray}{// \textbf{Adaptated Robot State}}$

        \State $\mathbf{\tilde{o}}_t \leftarrow \mathcal{S}^{}_{\text{sce}} \times \mathcal{S}'_{\text{rob}}$ 
        
        \State $\mathbf{a}^{K}_t \sim \mathcal{N}(0, \mathbf{I})$ \textcolor{gray}{// \textbf{Diffusion Process:}}
        \Repeat
            \State $k \leftarrow K-1$, and $K \leftarrow K-1$
            \State $\mathbf{a}^{k}_t \leftarrow \mathcal{N}\big(\mu_{k}(\mathbf{a}^{k+1}_{t}, \epsilon_{\theta}(\mathbf{a}^{k+1}_{t},\mathbf{\tilde{o}}_t,k+1)), 0\big)$
            \State \textbf{if} $k \leq 5$: \textcolor{gray}{// \textbf{Multi-objective Projection:}}
            \State\quad\ \textbf{for} $j \leftarrow 0$ \textbf{to} $T_a-1$ \textbf{do}
            \State\quad\quad\ \ $\nu^{k*}_{t+j} \leftarrow \arg\min\limits_{\nu} \|\nu^k_{t+j}\|^2_2$
            \State\quad\quad\ \ $\text{s.t. }$ Cumulative constraints \eqref{eq:sequence_constraint}, \eqref{eq:sequence_constraint-2}
            \State\quad\ $\mathbf{a}^{k}_{t:t+T_a-1} \leftarrow \Phi^{-}[\Phi(\mathbf{a}^{k}_{t:t+T_a-1}) + \nu^{k*}_{t:t+T_a-1}]$
        \Until{$K=0$}
        \State $\tau \leftarrow \text{Decode}(\mathbf{a}^{0}_{t:t+T_a-1})$
    \end{algorithmic}
\end{algorithm} 
\vspace{-0.5em}

The projected actions $\mathbf{a}^{k*}_t = \Phi^{-}[\Phi(\mathbf{a}^{k}_t) + \nu^{k*}_t]$ guarantee:
\begin{equation}
\mathbb{P}\left(\bigcap_{j=0}^{T_a} \{\mathcal{S}'_{\text{rob}}(z)_{t+j} \geq \epsilon_{\text{safe}}\}\right) = 1,
\end{equation} 
indicating the cumulative trajectory is always safe, with an example of $\mathcal{S}'_{\text{rob}}(z)_{t+1}=\mathcal{S}'_{\text{rob}}(z)_t+\Phi(\mathbf{a}^{k*}_t)$.

\section{Experiments}

We evaluate the effectiveness of our adaptation-projection strategy in improving the applicability of the trained diffusion policies for pick-and-place, pushing, and pouring tasks across manipulator configurations and task requirements. Each task's demonstrations are collected with a Franka robot equipped with two gripper configurations (the Franka gripper with flexible fingertips or Robotiq-2f gripper) at 2Hz. For example, the pick-and-place and pushing tasks are performed with the former gripper, while the pouring task is performed with the latter. Each task policy is trained with around 60 demonstrations, and the training scheme is consistent with the Simple DP3 framework~\cite{Ze2024DP3} without any modifications.

\subsection{Pick-and-Place Task}
We first evaluate the policy's performance in pick-and-place tasks with cross-gripper and cross-manipulator setups.

(a) \textbf{Baselines}: Diffusion Policy (DP)~\cite{chi2023diffusion},  Diffusion Policy 3D (DP 3D)~\cite{Ze2024DP3}, BC-z\cite{jang2022bc}, Ours w/o AP (Adaptation-Projection strategy), and Ours w/o $\mathcal{G}^*_{\text{prob}}$. (b) \textbf{Setup}: Our policy and baselines are evaluated in different task settings, including different grippers, different objects, and placing the object to different-height platforms. Notably, all training demonstrations included the base gripper ($\mathbb{G}_0$), a single object (a pink block), and a fixed-height platform (9 cm). In this task, the visual observation contains global-scenario point clouds acquired via a Realsense L515 camera, and a ego-camera Kinect Azure to collect RGB-D images. A failure trial refers to failing to grasp the object, or dropping it during manipulation, or encountering collisions.

\textbf{Incorporate gripper-agnostic grasping knowledge:} To investigate how switching grippers with different morphologies in the policy's inference stage affect the policy's performance, we incorporate gripper-agnostic grasping knowledge during policy training as a comparison. Specifically, we introduce a gripper-agnostic \textit{grasping probability map} $\mathcal{G}_{\text{prob}}$ as an additional observation component. This map captures object-centric grasp affordances that are independent of end-effector geometry, guiding the policy to focus on relevant object features rather than gripper-specific visual patterns. By decoupling object-related cues from the gripper's appearance, $\mathcal{G}_{\text{prob}}$ enhances the policy's robustness to variations in gripper morphology. We adopt the Generative Grasping CNN (GG-CNN) \cite{morrison2020learning} for $\mathcal{G}_{\text{prob}}$ synthesis from depth images. However, real-world pick-and-place manipulations introduce two key challenges: (1) the hand-eye camera moving with the robot, causing scale variations in object pixels, and (2) lighting changes disturb depth sensor readings. These factors degrade GG-CNN's output stability, i.e., $\mathcal{G}_{\text{prob}}$, and destabilize policy training and inference performance. To address this issue, our solution involves: (1) threshold filtering: discard pixels with $\mathcal{G}_{\text{prob}} < 0.7$, (2) centroid computation: $\mathds{O} = \frac{1}{N}\sum_{i=1}^N (u_i,v_i)$ for remaining pixels, and (3) region masking: generate $\mathcal{G}^*_{\text{prob}}$ through circular masking ($r=30$ pixels) about $\mathds{O}$. The map satisfies $\mathcal{G}^*_{\text{prob}}(u,v)=1 \text{, if } \|(u,v) - \mathds{O}\|_2 \leq 30$. This spatial filtering maintains grasp affordance information while eliminating outlier predictions caused by sensor noise, as shown in Fig.\ref{fig:gmap}. The policy observation consists of $\mathcal{O}^* = \mathcal{G}^*_{\text{prob}}\times \mathcal{S}_{\text{sce}}\times \mathcal{S}'_{\text{rot}}$.

\input{figure-gmap-v2}
\begin{table}[!t]
  \begin{center}
    \caption{Pick-and-place task across grippers (unit: \%).}
    \label{tab:comparison}
    \resizebox{\columnwidth}{!}{
    \begin{threeparttable}
        \begin{tabular}{l|c|c|c|c|c|c|c} 
            \toprule[1.5pt]
            \midrule[0.2pt]
            \multicolumn{8}{c}{Case 1: \textcolor{cyan}{Robot:} Franka (seen), \textcolor{Salmon}{Object:} {Block (seen)}, \textcolor{blue}{Platform height:} 9cm}\\
            \midrule
           Method & $\mathbb{G}_0$ & $\mathbb{G}_1$ & $\mathbb{G}_2$ & $\mathbb{G}_3$ & $\mathbb{G}_4$ & $\mathbb{G}_5$ & $\mathbb{G}_6$ \\
          \midrule
          Diffusion Policy & 20.0 & 0.0 & 60.0 & 40.0 & 0.0 & 40.0 & - \\
          Diffusion Policy 3D & 20.0 & 0.0 & 60.0 & 60.0 & 0.0 & 0.0 & -\\
          DP + AP & 100.0 & - & - & - & - & - & -\\
          DP 3D + AP & 80.0 & - & - & - & - & - & -\\
          BC-z & 20.0 & 0.0 & 20.0 & 20.0 & 0.0 & 0.0 & -\\
          Ours \textcolor{red}{w/o} AP & 100.0 & 0.0 & 60.0 & 40.0 & 0.0 & 0.0 & - \\
          Ours \textcolor{red}{w/o} $\mathcal{G}^*_{\text{prob}}$ & 80.0 & 20.0 & 40.0 & 80.0 & 100.0 & 0.0 & - \\
          Ours & \textbf{100.0} & \textbf{80.0} & \textbf{100.0} & \textbf{80.0} & \textbf{100.0} & \textbf{100.0} & \textbf{-}\\

        \midrule[0.2pt]
        
        \multicolumn{8}{c}{Case 2: \textcolor{cyan}{Robot:} Franka (seen), \textcolor{Salmon}{Object:} {Banana (\textcolor{red}{unseen})}, \textcolor{blue}{Platform height:} 9cm}\\
        \midrule
          Diffusion Policy & 20.0 & 0.0 & 40.0 & \textbf{60.0} & 40.0 & 20.0 & -\\
          Diffusion Policy 3D & 20.0 & 0.0 & 40.0 & 40.0 & 20.0 & 0.0 & -\\
          Ours \textcolor{red}{w/o} AP & 80.0 & 0.0 & 40.0 & 40.0 & 0.0 & 20.0 & - \\
          Ours \textcolor{red}{w/o} $\mathcal{G}^*_{\text{prob}}$ & 60.0 & 0.0 & 40.0 & \textbf{60.0} & 40.0 & 0.0 & - \\
          Ours & \textbf{80.0} & \textbf{60.0} & \textbf{100.0} & \textbf{60.0} & \textbf{60.0} & \textbf{60.0} & \textbf{-}\\
        \midrule[0.2pt]

        \multicolumn{8}{c}{Case 3: \textcolor{cyan}{Robot:} Kuka (\textcolor{red}{unseen}), \textcolor{Salmon}{Object:} {Block (seen)}, \textcolor{blue}{Platform height:} \textcolor{red}{22cm}}\\
        \midrule
          Diffusion Policy & - & - & - & - & - & 0.0 & 0.0\\
          Diffusion Policy 3D & - & - & - & - & - & 0.0 & 0.0\\
          Ours & - & - & - & - & - & \textbf{80.0} & \textbf{80.0}\\
        \midrule[0.2pt]
        
        \bottomrule[1.5pt]
        \end{tabular}
    
        \begin{tablenotes}[para,flushleft]
            \small
            - The training data is collected with the base gripper \textcolor{red}{$\mathbb{G}_0$} and a pink block. \\
            - The initial pose and position of the object are identical across tests.\\
            - \textcolor{red}{$\mathbb{G}_i$} indicates different grippers reported in Fig. \ref{fig:grippers}. \textcolor{red}{w/o}: without this module.
            - Every method is validated 5 times for each gripper, totaling 30 evaluations. The Robotiq-3f gripper ($\mathbb{G}_6$) is too heavy to equip on Franka robot, thus $\mathbb{G}_6$ is evaluated on Kuka robot.

        \end{tablenotes}
    \end{threeparttable}
    }
  \end{center}
  \vspace{-1.4em}
\end{table}

Table \ref{tab:comparison} summarizes the success rates of different policies across-manipulator configurations (e.g., robot and gripper types) and different task requirements (e.g., object types and placement heights). Here, $\mathbb{G}_i$ represents different grippers in Fig. \ref{fig:grippers} and $\mathbb{G}_0$ is the base gripper. The results showcase the following trends: (1) \textbf{Our full method} consistently outperforms the baselines across all different manipulator configurations, demonstrating its effectiveness in adapting to different manipulator and task setups. (2) \textbf{Baseline policies (DP, DP 3D, BC-z)} perform poorly, often generating imprecise trajectories that lead to collisions for longer gripper ($\mathbb{G}_1$) and unable grasping for short one ($\mathbb{G}_4$). Their performance is highly sensitive to the specific gripper geometry and they fail to adapt to unseen objects, especially when object positions are randomized. (3) \textbf{The ablation study without the Adaptation-Projection strategy (Ours w/o AP)} fails when using grippers with geometries that significantly differ from the base gripper. This highlights the critical role of the projection module in adapting to physical hardware changes. (4) \textbf{The ablation study without gripper-agnostic knowledge (Ours w/o $\mathcal{G}^*_{\text{prob}}$)} shows a significant performance drop when grippers appear visually different from the ego-centric camera's view or when handling unseen objects. This underscores the importance of decoupling grasp-relevant features from the specific gripper's appearance for robust generalization. (5) \textbf{Adapting to New Task Requirements:} Our method seamlessly adapts to new task requirements, such as the varying platform heights in Case 3. We encode these changes directly into our adaptation process by treating the height difference as a TCP shift and applying the translational adaptation from \eqref{eq:gripper_mapping}. This allows our method to maintain high performance, even with challenging grippers like $\mathbb{G}_6$, whereas baseline methods completely fail to adapt to these changes. Fig. \ref{fig:tracking_rollout} visualizes a successful rollout under these modified task and manipulator configurations.

\input{figure-exp-horizon-small}

\subsection{Pushing task}

\begin{figure}[!t]
    \centering
    \begin{tikzpicture}
        \node(border)[minimum height=2.5cm, minimum width=8.5cm, draw=none, dashed] at (0,0){};
        \node[inner sep=0pt,anchor=west](data)at($(border.west)+(0,0)$) {\includegraphics[width=0.48\textwidth]{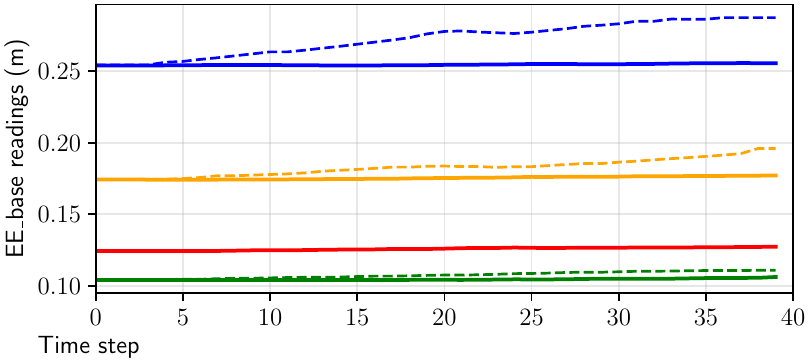}};

        \node(g5)[font=\footnotesize, anchor=south, color=red] at ($(data.south)+(-1,1.15)$) {Franka + gripper $\mathbb{G}_5$};

        \node(g1)[font=\footnotesize, anchor=west, color=orange] at ($(g5.east)+(1,0.3)$) {Franka + gripper $\mathbb{G}_1$};

        \node(g6)[font=\footnotesize, anchor=north, color=blue] at ($(data.north)+(0.5,-0.7)$) {Kuka + gripper $\mathbb{G}_6$};

        \draw[line width=1pt, black] ($(data.south)+(-2,0.15)$) -- ++ (0.5,0) node(label1)[right]{\footnotesize refined trajectory};
        \draw[color=black,line width=1pt, dashed] ($(label1.east)+(0.3,0)$) -- ++ (0.5,0.) node[right]{\footnotesize unrefined trajectory};

    \end{tikzpicture}
    \caption{Result of the pushing task in different configurations.}
    \label{fig:pushing_result}
    \vspace{-1.3em}
\end{figure}

\input{figure-exp-horizon-pour}

\begin{figure}[!t]
    \centering
    \begin{tikzpicture}
        \node(border)[minimum height=2.5cm, minimum width=8.5cm, draw=none, dashed] at (0,0){};
        \node[inner sep=0pt,anchor=west](data)at($(border.west)+(0,0)$) {\includegraphics[width=0.48\textwidth]{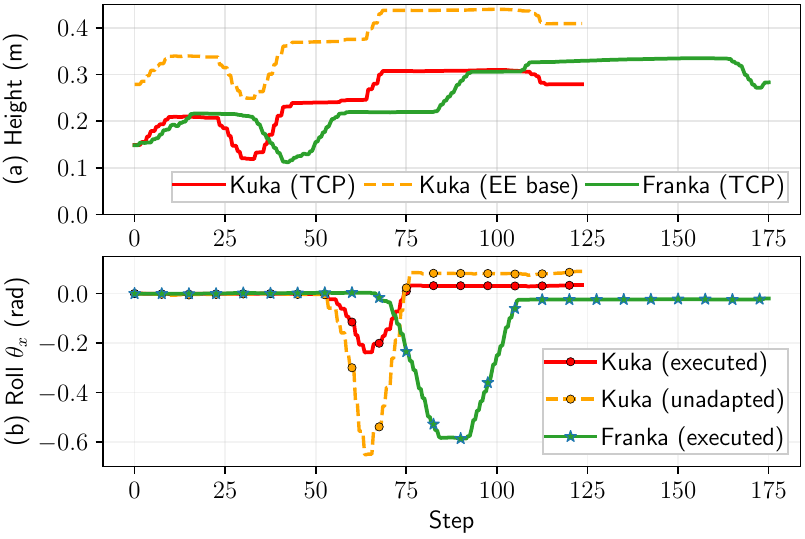}};

        \node(t1)[color=red,anchor=south] at ($(data.north)+(2.2,-1.8)$){6};
        \draw[line width=1pt,draw=none,fill=cyan!30!white] (t1.center) circle (5.5pt);
        \node[color=red] at (t1.center){6};
        \draw[-stealth,color=black,line width=1pt] ($(t1.north)$) -- ++ (-0.23,0.36);

        \node(t2)[color=red,anchor=north] at ($(t1.north)+(1.7,0)$){6};
        \draw[line width=1pt,draw=none,fill=cyan!30!white] (t2.center) circle (5.5pt);
        \node[color=forestgreen] at (t2.center){6};
        \draw[-stealth,color=black,line width=1pt] ($(t2.north)$) -- ++ (0,0.36);

        \node(tt1)[color=red,anchor=south] at ($(data.south)+(-1.2,1.75)$){4};
        \draw[line width=1pt,draw=none,fill=cyan!30!white] (tt1.center) circle (5.5pt);
        \node[color=red] at (tt1.center){4};
        \draw[-stealth,color=black,line width=1pt] ($(tt1.east)$) -- ++ (0.53,0);

        \node(tt2)[color=red,anchor=south] at ($(tt1.south)+(1.72,-0.1)$){4};
        \draw[line width=1pt,draw=none,fill=cyan!30!white] (tt2.center) circle (5.5pt);
        \node[color=forestgreen] at (tt2.center){4};
        \draw[-stealth,color=black,line width=1pt] ($(tt2.south)$) -- ++ (0,-0.53);

    \end{tikzpicture}
    \caption{Result of the pouring task in different manipulators.}
    \label{fig:pouring_task_result}
    \vspace{-1.3em}
\end{figure}

To isolate and evaluate the effect of our adaptation-projection strategy on geometric variations alone, we design a pushing task in which the policy relies exclusively on low-dimensional robot states, without incorporating any visual input. This configuration eliminates confounding effects from visual changes and directly tests the policy's capacity to accommodate various manipulator morphologies. In this task, the robot is required to push an object forward (with the object initially placed on the table and grasped by the gripper). Without our strategy, switching between different manipulator configurations causes variations in the EE\_base readings, leading to out-of-distribution issues in the robot's action-state distribution and resulting in unintended behaviors, such as lifting the object during the pushing motion (see Fig.\ref{fig:pushing_result}). In contrast, our strategy successfully refines the generative trajectories across configurations, ensuring reliable task execution.

\subsection{Pouring tea task}

Pouring task is designed to evaluate our method on a more complex, multi-stage manipulation sequence. This requires the robot to pick up a teapot, pour water into a cup with precision, and then place the teapot back on a platform. This task is particularly challenging as it demands accurate control of both position and orientation throughout the entire trajectory. Our adaptation-projection strategy is critical for dynamically refining the generated trajectory to account for TCP shifts between the base and new robot configurations in both translation \eqref{eq:gripper_mapping} and rotation \eqref{eq:rotational_mapping}. This ensures collision-free motion during the pick-and-place phases and enables accurate, spill-free pouring. Fig. \ref{fig:pouring_rollout} demonstrates rollout of pouring in different manipulators, and Fig. \ref{fig:pouring_task_result} shows the corresponding trajectories in different axes. The results demonstrate that our method successfully adapts the trained policy to different manipulators, refining the trajectory in translational ($z$-axis in Fig. \ref{fig:pouring_task_result}(a)) and rotational ($\theta_x$ in Fig. \ref{fig:pouring_task_result}(b)) axes and ensuring task completions in pouring tasks across various configurations.

\section{CONCLUSIONS AND OUTLOOK}

This paper introduces an inference-stage adaptation-projection strategy for diffusion policy, transferring manipulation skills across different grippers and robots. This transition does not require retraining or fine-tuning the policy with the new manipulator configuration. Instead, it only needs to introduce the configuration or new task requirements in the policy inference phase, refining the generated trajectories to satisfy safety constraints and task completion. This approach effectively reduces the time and cost of data collection and model training for each new manipulator or task. We validate our method on various manipulation tasks, including pick-and-place, pushing, and pouring, using different robots (typically including the Franka Panda and Kuka iiwa 14) and parallel grippers (e.g., Robotiq 2F/3F grippers, flexible fingertips, 3D-printed designs). The results demonstrate that our method achieves high success rates in these cross-manipulator scenarios, showcasing its effectiveness and practicality. This approach can be extended to other manipulation tasks and gripper types, such as deformable object manipulation (e.g., cloth manipulation) and multi-fingered hands. Future work could explore more complex task requirements and constraints, such as force control and compliance, to further enhance the adaptability and robustness of diffusion policies in real-world applications.

\normalem
\bibliographystyle{./IEEEtran} 
\bibliography{./IEEEexample}

\end{document}